\documentclass[final]{cvpr}

\usepackage{times}
\usepackage{epsfig}
\usepackage{graphicx}
\usepackage{amsmath}
\usepackage{amssymb}
\usepackage{float}
\usepackage{booktabs}
\usepackage{subfigure}
\usepackage{hyperref}
\usepackage{multirow}
\usepackage{url}
\usepackage{caption}

\newcommand{\Tref}[1]{Table~\ref{#1}}
\newcommand{\Eref}[1]{Equation~\ref{#1}}
\newcommand{\Fref}[1]{Figure~\ref{#1}}
\newcommand{\Sref}[1]{Section~\ref{#1}}

\usepackage[marginal]{footmisc}

\def\confYear{CVPR 2021}

\begin{document}
\title{Multi-attentional Deepfake Detection}

\author{Hanqing Zhao\textsuperscript{\rm 1}\qquad  Wenbo Zhou\textsuperscript{\rm 1,}\textsuperscript{$\dagger$} \qquad Dongdong Chen\textsuperscript{\rm 2}\\
Tianyi Wei\textsuperscript{\rm 1} \qquad Weiming Zhang\textsuperscript{\rm 1,}\textsuperscript{$\dagger$}\qquad Nenghai Yu\textsuperscript{\rm 1}\\ University of Science and Technology of China\textsuperscript{\rm 1} \qquad Microsoft Cloud AI\textsuperscript{\rm 2}\\
 {\tt\small {\{zhq2015@mail, welbeckz@, bestwty@mail, zhangwm@, ynh@\}.ustc.edu.cn}} \\ 
{\tt\small {cddlyf@gmail.com}}}

\maketitle

\footnote{\textsuperscript{$\dagger$}\  Corresponding Author.}

\maketitle

\begin{abstract}
 Face forgery by deepfake is widely spread over the internet and has raised severe societal concerns. Recently, how to detect such forgery contents has become a hot research topic and many deepfake detection methods have been proposed. Most of them model deepfake detection as a vanilla binary classification problem, i.e, first use a backbone network to extract a global feature and then feed it into a binary classifier (real/fake). But since the difference between the real and fake images in this task is often subtle and local, we argue this vanilla solution is not optimal. In this paper, we instead formulate deepfake detection as a fine-grained classification problem and propose a new multi-attentional deepfake detection network. Specifically, it consists of three key components: 1) multiple spatial attention heads to make the network attend to different local parts; 2) textural feature enhancement block to zoom in the subtle artifacts in shallow features; 3) aggregate the low-level textural feature and high-level semantic features guided by the attention maps. Moreover, to address the learning difficulty of this network, we further introduce a new regional independence loss and an attention guided data augmentation strategy. Through extensive experiments on different datasets, we demonstrate the superiority of our method over the vanilla binary classifier counterparts, and achieve state-of-the-art performance. The models will be released recently at \url{https://github.com/yoctta/multiple-attention}.

\end{abstract}

\section{Introduction}
Benefiting from the great progress in generative models, deepfake techniques have achieved significant success recently and various face forgery methods \cite{2016transfiguring,thies2016face2face:,koujan2020head2head,nirkin2019fsgan:,pumarola2018ganimation:,wu2018reenactgan:,Natsume2018RSGANFS,lipsync} have been proposed. As such techniques can generate high-quality fake videos that are even indistinguishable for human eyes, they can easily be abused by malicious users to cause severe societal problems or political threats. 
To mitigate such risks,many deepfake detection approaches \cite{matern2019exploiting,rossler2019faceforensics++,li2020face,shao2020thinking,masi2020two,wu2020sstnet} have been proposed. Most of them model deepfake detection as a vanilla binary classification problem (real/fake). Basically, they often first use a backbone network to extract global features of the suspect image and then feed them into a binary classifier to discriminate the real and fake ones.  

\begin{figure}[t]
    \begin{center}
        \includegraphics[width=0.75\linewidth]{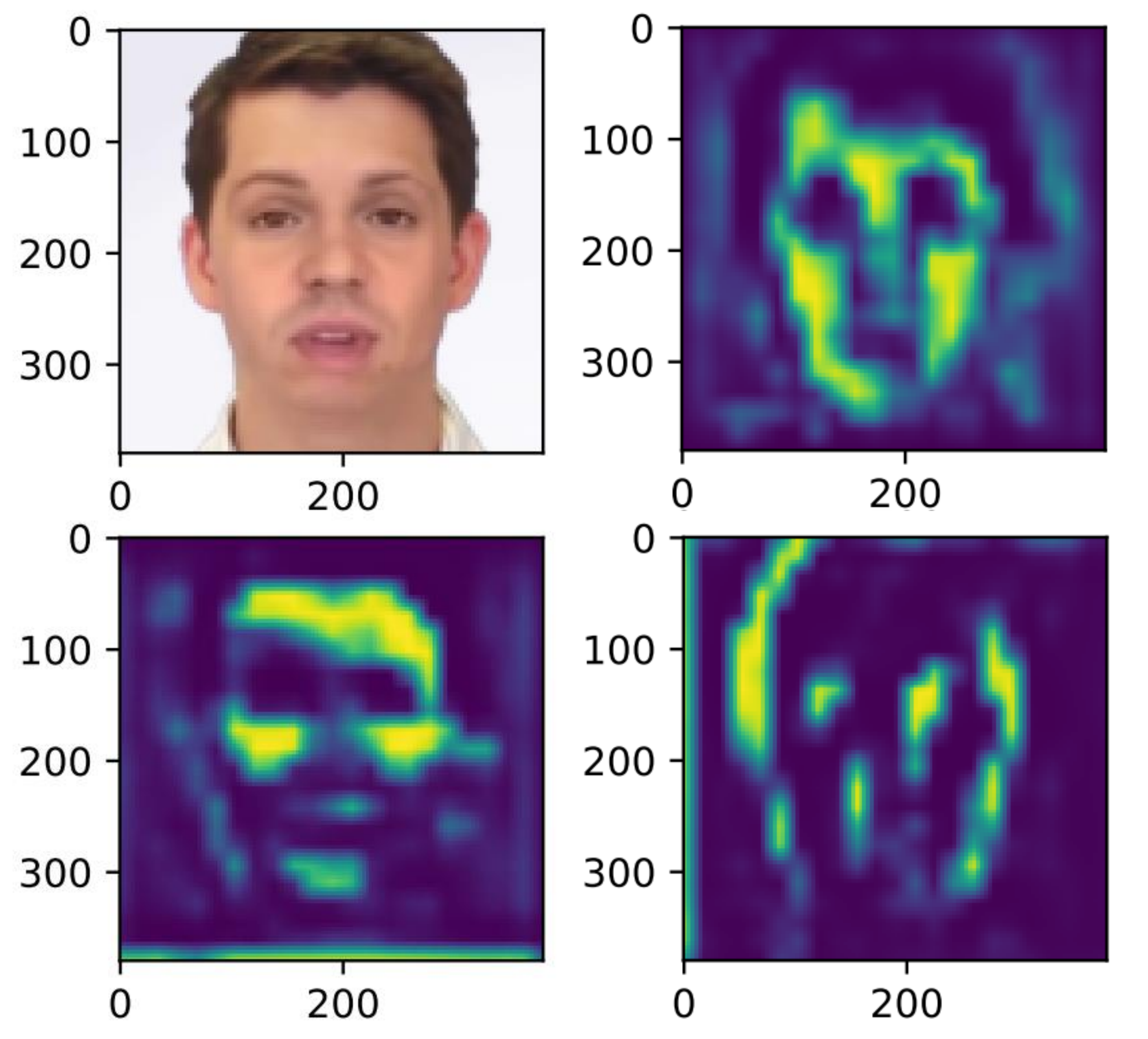}
    \end{center}
    \vspace{-1.8em}
    \caption{Example of the multiple attentional regions obtained by our method. The attention regions are separated and respond to different discriminative features.}
    \label{fig_attentionmaps}
\end{figure}

However, as the counterfeits become more and more realistic, the differences between real and fake ones will become more subtle and local, thus making such global feature based vanilla solutions work not well. But actually, such subtle and local property shares a similar spirit as the fine-grained classification problem. For example, in the fine-grained bird classification task, some species look very similar and only differentiate from each other by some small and local differences, such as the shape and color of the beak. Based on this observation, we propose to model deepfake detection as a special fine-grained classification problem with two categories. 

Inspired by the success of parts based model in the fine-grained classification field, this paper presents a novel multi-attention network for deepfake detection. First, in order to make the network attend to different potential artifacts regions, we design multi-attention heads to predict multiple spatial attention maps by using the deep semantic features. Second, to prevent the subtle difference from disappearing in the deep layers, we enhance the textural feature obtained from shallow layers and then aggregate both low-level texture features and high-level semantic features as the representation for each local part. Finally, the feature representations of each local part will be independently pooled by a bilinear attention pooling layer and fused as the representation for the whole image. \Fref{fig_attentionmaps} gives an example of the discriminative features obtained by our method.

However, training such a multi-attentional network is not a trivial problem. This is mainly because that, unlike single-attentional network \cite{dang2020detection} which can use the video-level labels as explicit guidance and be trained in a supervised way, the multi-attentional structure can only be trained in an unsupervised or weakly-supervised way. By using a common learning strategy, we find the multi-attention heads will degrade to a single-attention counterpart, i.e.,  only one attention region produces a strong response while all remaining attention regions are suppressed and can not capture useful information. To address this problem, we further propose a new attention guided data augmentation mechanism. In detail, during training, we will deliberately blur some high-response attention region (\textbf{soft attention dropping}) and force the network to learn from other attention regions. Simultaneously, we introduce a new regional independence loss to encourage different attention heads to attend to different local parts. 

To demonstrate the effectiveness of our multi-attentional network, we conduct extensive experiments on different existing datasets, including FaceForensics++\cite{rossler2019faceforensics++}, Celeb-DF\cite{li2020celeb} and DFDC\cite{dolhansky2020deepfake}. It shows that our method is superior to the vanilla binary classifier baselines and achieves state-of-the-art performance. 
In summary, the contributions of this paper are threefold as below:

\begin{itemize}
  \item We reformulate the deepfake detection as a fine-grained classification task, which brings a novel perspective for this field.
  
  \item We propose a new multi-attentional network architecture to capture local discriminative features from multiple face attentive regions. To train this network, we also introduce a regional independence loss and design an attention guided data augmentation mechanism to assist the network training in an adversarial learning way.
  
  \item Extensive experiments demonstrate that our method outperforms the vanilla binary classification baselines and achieves state-of-the-art detection performance.

\end{itemize}

\section{Related Works}\label{sec_2}

Face forgery detection is a classical problem in computer vision and graphics. Recently, the rapid progress in deep generative models makes the face forgery technique ``deep" and can generate realistic results, which presents a new problem of deepfake detection and brings significant challenges. Most deepfake detection methods solve the problem as a vanilla binary classification, however, the subtle and local modifications of forgeried faces make it more similar to fine-grained visual classification problem.

\subsection{Deepfake Detection}\label{sec_2.1}
Since the face forgery causes great threat to societal security, it is of paramount importance to develop effective countermeasures against it. Many works \cite{headpose,eyeblink,ciftci2020fakecatcher,Zhou2017TwoStreamNN,rossler2019faceforensics++,li2020face,shao2020thinking,masi2020two,wu2020sstnet,wang2019detecting} have been proposed. Early works \cite{headpose,eyeblink} detect the forgery through visual biological artifacts, e.g., unnatural eye blinking or inconsistent head pose. 

As the learning based methods become mainstream, some works \cite{Zhou2017TwoStreamNN,rossler2019faceforensics++} have proposed frameworks which extract features from spatial domain and have achieved excellent performances on specific datasets. Recently, more data domains have been considered by emerging methods. \cite{wu2020sstnet} detects tampered faces through Spatial, Steganalysis and Temporal features. It adds a stream of simplified Xception with a constrained convolution layer and an LSTM. \cite{masi2020two} uses a two-branch representation extractor to combine information from the color domain and the frequency domain using a multi-scale Laplacian of Gaussian (LoG) operator. \cite{shao2020thinking} uses frequency-aware decomposition and local frequency statistic to expose deepfake artifacts in frequency domain and achieves state-of-the-art performance.

Most existing methods treat the deepfake detection as a universal binary classification problem. They focus on how to construct sophisticated feature extractors and then a dichotomy to distinguish the real and fake faces. However, the photo-realistic counterfeits bring significant challenge to this binary classification framework. In this paper, we redefine the deepfake detection problem as a fine-grained classification problem according to their similarity.

\begin{figure*}[t]
    \begin{center}
        \includegraphics[width=0.9\linewidth]{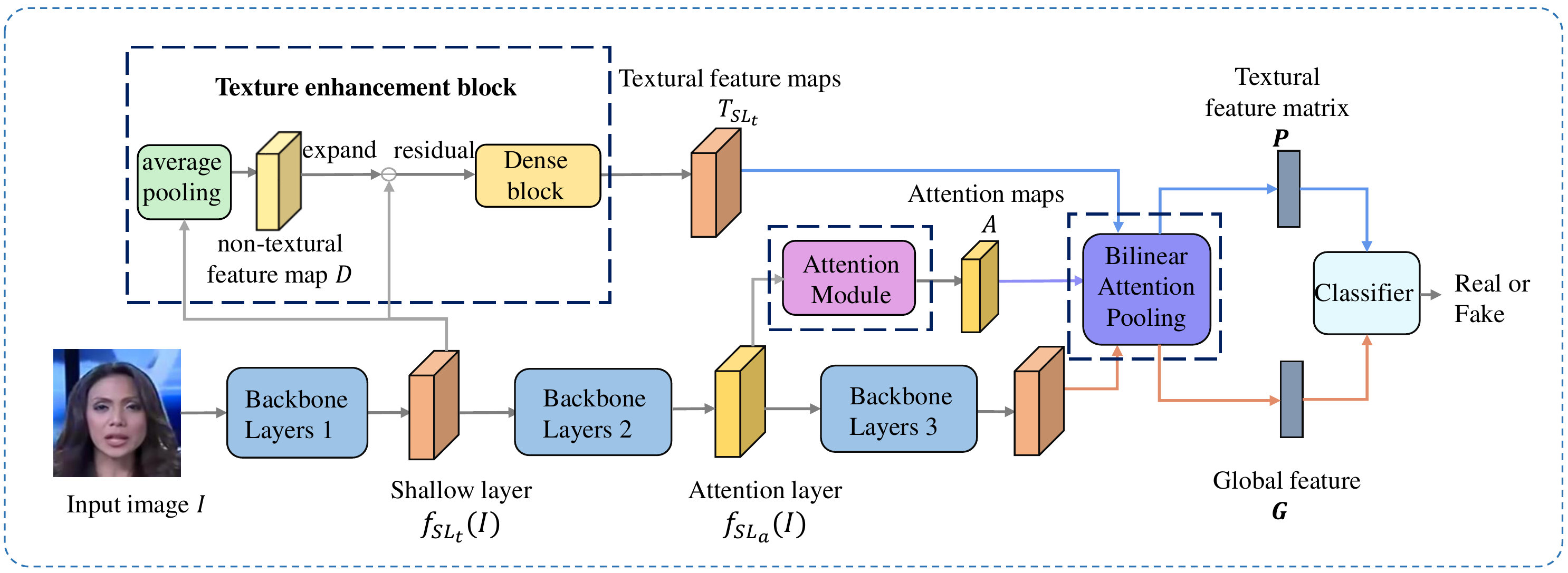}
    \end{center}
    \vspace{-2em}
    \caption{\small The framework of our method. Three components play an important role in our framework: an Attention Module for generating multiple attention maps, a texture enhancement block for extracting and enhancing the textural information and a bidirectionally used bilinear attention pooling for aggregating textural and semantic features. }
    \label{fig_architecture}
\end{figure*}

\subsection{Fine-grained Classification}\label{sec_2.2}
Fine-grained classification \cite{deform13,partrcnn,rcnn2014,2015part,racnn,macnn,Yang2018Learning,wsdan,du2020fine} is a challenging research task in computer vision, which captures the local discriminative features to distinguish different fine-grained categories. Studies in this field mainly focus on locating the discriminative regions and learning a diverse collection of complementary parts in weakly-supervised manners. Previous works \cite{deform13,partrcnn} build part models to localize objects and treat the objects and semantic parts equally. Recently, several works \cite{macnn,Yang2018Learning,du2020fine} have been proposed under a multiple attentional framework, the core ideal of these method is that learning discriminative regions in multiple scales or image parts simultaneously and encouraging the fusion of these features from different regions. In addition, \cite{wsdan} designs attention cropping and attention dropping to obtain more balanced attention maps. In this paper, we model deepfake detection as a special fine-grained classification problem for the first time. It shares the same spirit in learning subtle and discriminative features, but only involves two categories, \ie, real and fake.

\section{Methods}

\subsection{Overview}\label{sec_3.1}
 In this section, we initially state the motivation of the designing and give a brief overview of our framework. As aforementioned, the discrepancy between real and fake faces is usually subtle and occurs in local regions, which is not easy to be captured by single-attentional structural networks. Thus we argue that decomposing the attention into multiple regions can be more efficient for collecting local feature for deepfake detection task. Meanwhile, the global average pooling which is commonly adopted by current deepfake detection approaches is replaced with local attention pooling in our framework. This is mainly because the textural patterns vary drastically among different regions, the extracted features from different regions may be averaged by the global pooling operation, resulting in a loss of distinguishability.  On the other hand, we observe that the slight artifacts caused by forgery methods tend to be preserved in textural information of shallow features. Here, the textural information  represents the high frequency component of the shallow features, just like the residual information of RGB images. Therefore, more shallow feature should be focused on and enhanced, which has not been considered by current state-of-the-art detection approaches.
 
 Motivated by these observation, we propose a multi-attentional framework to solve the deepfake detection as a fine-grained classification problem. In our framework, three key components are integrated into the backbone network: 1) We employ an attention module to generate multiple attention maps. 2) We use densely connected convolutional layers \cite{densenet} as a texture enhancement block, which can extract and enhance the textural information from shallow feature maps. 3) We replace the global average pooling with Bilinear Attention Pooling(\textbf{BAP}). And we use \textbf{BAP} to collect the textural feature matrix from the shallow layer and retain the semantic feature from the deep layer. The framework of our method is depicted in \Fref{fig_architecture}.

\begin{figure}[t]
    \begin{center}
        \includegraphics[width=0.9\linewidth]{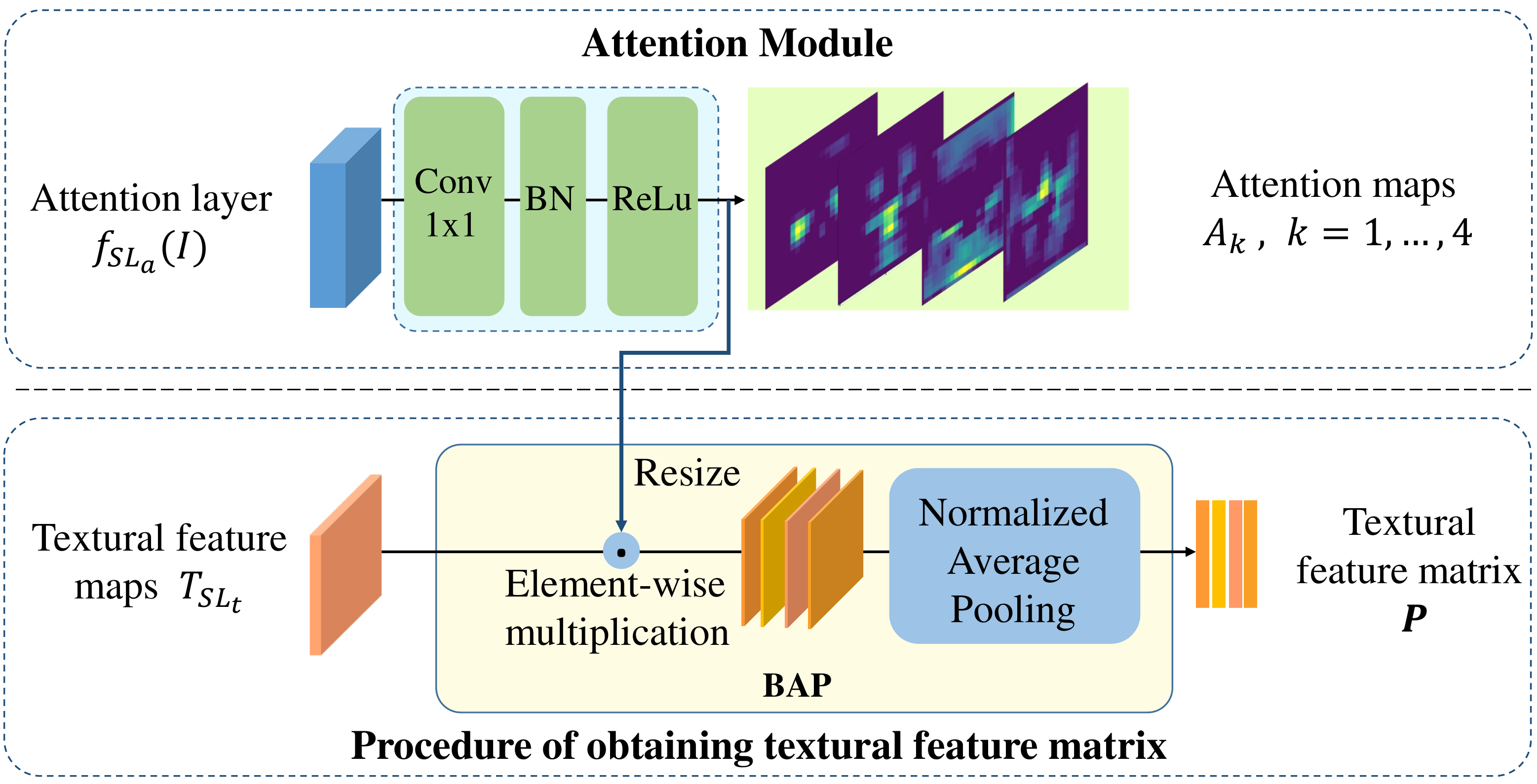}
    \end{center}
    \vspace{-1em}
    \caption{\small The structure of attention module and the procedure of obtaining textural feature matrix $\textbf{P}$. The proposed normalized average pooling is adopted instead of global average pooling.}
    \label{fig_multi_attention}
\end{figure}

Unlike single-attentional structure based network which can take the video-level labels as explicit guidance for training, the multi-attentional based network can only be trained in a unsupervised or weakly-supervised manner due to the lack of region-level labels. It could lead to a degradation of network that multiple attention maps focus on same region while ignoring other regions which may also provide discriminative information. To address the problem, we specifically design a Region Independence Loss, which aims to ensure each attention map focusing on one specific region without overlapping and the focused region is consistent across different samples. Further, we employ the Attention Guided Data Augementation(\textbf{AGDA}) mechanism to decrease the salience of the most discriminative feature and force other attention maps to mine more useful information.

\subsection{Multi-attentional Framework}\label{sec_3.2}

Denote the input face image of network as $I$ and the backbone network of our framework as $f$, the feature maps extracted from the intermediate stage of $t$-th layer is denoted as $f_t(I)$ with size of $C_t\times H_t\times W_t$. Here, $C_t$ is the number of channels, $H_t, W_t$ are the height and the width of feature maps, respectively.

\vspace{0.1em}
\noindent\textbf{Multiple Attention Maps Generation.} As described above, given a real/fake face image $I$ as input, our framework first uses an attention block to generate multiple attention maps for $I$. As shown in \Fref{fig_multi_attention}, the attention block is a light weighted model which consists of a 1$\times$ 1 convolutional layer, a batch normalization layer and non-linear activation layer ReLU. The feature map extracted from a specific layer $SL_a$ will be fed into this attention block to obtain $M$ attention maps $A$ with size of $H_t\times W_t$, among which $A_k \in R^{\{H_t\times W_t\}}$ represents the $k$-th attention map and corresponds to one specific discriminative region, for example, eyes, mouth or even blending boundary defined in \cite{li2020face}. The determination of $SL_a$ will be discussed in \Sref{sec_4}.

\vspace{0.1em}
\noindent\textbf{Textural Feature Enhancement.}
Most binary classification frameworks of deepfake detection do not pay attention to an important phenomenon, that is, the artifacts caused by forgery methods are usually salient in the textural information of shallow feature maps. The textural information here represents the high frequency component of the shallow features. Thus to preserve more textural information for capturing those artifacts, we design a textural feature enhancement block as shown in \Fref{fig_multi_attention}. We first apply the local average pooling in patches to down-sample the feature maps from a specific layer $SL_t$ and obtain the pooled feature map $D$. How to choose $SL_t$ will be discussed in the following experiments part. Then similar to the texture representation of spatial image, we define the residual at the feature level to represent the texture information as below:
\begin{equation}
    T_{SL_t}=f_{SL_t}(I)-D
\label{eq_res}
\end{equation}

Here $T$ contains most textural information of $f_{SL_t}(I)$. We then use a densely connected convolution block with 3 layers to enhance $T$, the output is noted as $F\in R^{C_F\times H_s \times W_s}$, which is defined as ``textual feature map".

\vspace{0.1em}
\noindent\textbf{Bilinear Attention Pooling.}
After getting the attention map $A$ and textural feature map $F$, we use Bilinear Attention Pooling (BAP) to obtain feature maps. We bidirectionally use BAP for both shallow feature maps and deep feature maps. As shown in \Fref{fig_multi_attention}, to extract shallow textural feature, we first use bilinear interpolation to resize the attention maps into the same scale with feature maps if they are not match. Then, we respectively element-wise multiply textural feature map $F$ by each attention map $A_k$ and obtain partial textural feature maps $F_k$.

To the end of this step,  the partial textural feature maps $F_k$ should be fed into classifier after global pooling. However, considering the differences among the different region range, if using the traditional global average pooling, the pooled feature vector will be influenced by the intensity of attention map, which violates the purpose of focusing on textural information. To address the problem, we design a normalized average pooling:

\begin{equation}
% \begin{aligned}
    v_k=\frac{\sum_{m=0}^{H_s-1}\sum_{n=0}^{W_s-1}F_{k,m,n}}{||\sum_{m=0}^{H_s-1}\sum_{n=0}^{W_s-1}F_{k,m,n}||_2}
% \end{aligned}
\label{eq_norm_pool}
\end{equation}

The normalized attention features $v_k\in R^{1 \times N}$ are then stacked together to obtain the textural feature matrix $\mathbf{P} \in R^{M \times C_F} $, which will be fed into the classifier.

As to deep features, we first splice each attention map to get a single channel attention map $A_{sum}$. Then we use BAP for $A_{sum}$ and the feature map from the last layer of network to get the global deep feature $\mathbf{G}$, which will also be fed into the classifier.

\subsection{Regional Independence Loss for Attention Maps Regularization}\label{sec_3.3}

As aforementioned, training a multiple attention network may easily fall into  a network degraded case due to the lack of fine-grained level labels. In details, different attention maps tend to focus on the same region as shown in \Fref{fig_bad_attention_maps} which is not conducive to the network to capture rich information for a given input. In addition, for different input images, we hope that the each attention map locates in fixed semantic region, for example, attention map $A_1$ focuses on eyes in different image, $A_2$ focuses on mouth. Therefore, the randomness of captured information by each attention map will be reduced.

To achieve these goals, we propose a Region Independence Loss which helps to reduce the overlap among attention maps and keep the consistency for different inputs. We apply BAP on the pooled feature map $D$ otained in \Sref{sec_3.2} to get a ``semantic feature vector" : $V\in R^{M\times N}$, and the Regional Independence Loss is defined as below by modifying the center loss in \cite{deepfisherfaces}:

\begin{equation}
 \begin{aligned}
\mathcal{L}_{RIL}= \sum_{i=1}^{B}\sum_{j=1}^{M}\max(\left\|V_j^i-c_{j}^{t}\right\|_{2}^{2}-m_{in}(y_i),0)+\\
  \sum_{i,j\in(M,M),i\neq j}\max \left(m_{out}-\left\|c_{i}^{t}-c_{j}^{t}\right\|_{2}^{2}, 0\right)
 \end{aligned}
\label{eq_RILoss}
\end{equation}
where $B$ is the batch size, M is number of attentions, $m_{in}$ represents the margin between feature and the corresponding feature center and is set as different values when $y_i$ is $0$ and $1$. $m_{out}$ is the margin between each feature center. $c\in R^{M\times N}$ are feature centers of $V$, it is defined as below and updated in each iteration:
\begin{equation}
    c^{t}=c^{t-1}-\alpha\left(c^{t-1}-\frac{1}{B}\sum_{i=1}^B V^i\right)
\label{eq_feature_center}
\end{equation}

Here $\alpha$ is the updating rate of feature centers, we decay $\alpha$ after each training epoch. The first part of $L_{RIL}$ is an intra-class loss that pulls $V$ close to feature center $c$, the second part is an inter-class loss that repels feature centers scattered. we optimize $c$ by calculating the gradient for $V$ in each batch. Considering that the patterns of texture in fake faces should be more diverse than real ones for fakes are generated by multiple methods, thus we restrict part features of fake faces in the neighborhood from the feature center of real ones but with larger margin. In this way, we give a larger margin in the intra-class for searching useful information in  fake faces.

For the objective function of our framework, we combine this Regional Independence Loss with the traditional cross entropy loss:

\begin{equation}
% \begin{aligned}
\mathcal{L}=\lambda_1*\mathcal{L}_{CE}+\lambda_2*\mathcal{L}_{RIL}
% \end{aligned}
\label{eq_loss}
\end{equation}

$\mathcal{L}_{CE}$ is a cross entropy loss, $\lambda_1$ and $\lambda_2$ are the balancing weights for these two terms. By default, we set $\lambda_1 = \lambda_2 = 1$ in our experiments.

\subsection{Attention Guided Data Augmentations}\label{sec_3.4}

Under the restraining of Regional Independence Loss, we reduce the overlap of different attention regions. However, although different attention regions can be well separated, the attention maps may still respond to the same discriminative features. For example, in \Fref{fig_overlapattentions}, the attention regions are not overlapped but they all strongly respond to the landmarks of input faces. To force the different attention maps to focus on different information, we propose the Attention Guided Data Augmentation (\textbf{AGDA}) mechanism.

For each training sample, one of the attention maps $A_k$ is randomly selected to guide the data augmentation process, and it is normalized as Augmentation Map $A^*_k\in R^{H\times W}$. Then we use Gaussian blur to generate a degraded image. Finally, we use $A^*_k$ as the weight of original image and degraded image:

\begin{equation}
I'=I\times(1-A^*_k)+I_d\times A^*_k
\label{eq_aug}
\end{equation}

Attention guided data augmentation helps to train the models in two aspects. Firstly, it can add blurry to some regions which ensure the model to learn more robust features from other regions. Alternatively, AGDA can erase the most saliently discriminative region by chance, which forces different attention maps focusing their response on different targets. Moreover, the AGDA mechanism can prevent a single attention region from expanding too much and encourage the attention blocks to explore various attention region dividing forms.

\begin{table}[t]
\centering
\setlength{\tabcolsep}{3mm}
{\begin{tabular}{c|c|c}
\toprule
Candidate of $SL_t$ & Candidate of $SL_a$ & ACC(\%)   \\ \midrule
L2            & L4              & 96.38 \\ \midrule
L2            & L5              & \textbf{97.26} \\ \midrule
L3            & L4              & 96.14 \\ \midrule
L3            & L5              & 96.81 \\ 
\bottomrule
\end{tabular}}
\vspace{-0.8em}
\caption{Performance of our methods based on different combination of $SL_t$ and $SL_a$.}\label{tab_sl_selection}
\end{table}

\section{Experiments}\label{sec_4}

In this section, we first explore the optimal settings for our proposed multi-attentional framework and then present extensive experimental results to demonstrate the effectiveness of our method.

\subsection{Implement Details}\label{sec_4.1}
For all real/fake video frames, we use a state-of-the-art face extractor RetinaFace\cite{Deng2019RetinaFaceSD} to detect faces and save the aligned facial images as inputs with a size of $380\times 380$. We set hyper-parameters $\alpha=0.05 $ in \Eref{eq_feature_center}, and decayed by 0.9 after each epoch. The inter-class margin $m_{out}$ in \Eref{eq_RILoss} is set to 0.2. The intra-class margin $m_{in}$ are set as 0.05 and 0.1 respectively for real and fake images. We choose quantity of attention maps $M$, $SL_a$ and $SL_t$ by experiments.  In AGDA we set the resize factor 0.3 and Gaussian blur $\sigma=7$. Our models are trained with Adam optimizer\cite{ADAM} with learning rate 0.001 and weight decay 1e-6. We train our models on 4 RTX 2080Ti GPUs with batch size 48.

\subsection{Determination of $SL_a$ and $SL_t$}\label{sec_4.2}
In this paper, we adopt EfficientNet-b4\cite{efficientnet} as the backbone network of our multi-attentional framework. EfficientNet-b4 is able to achieve comparable performance to XceptionNet \cite{xception} with only half FLOPs. There are 7 main layers in total of EfficientNet, which are denoted from L1-L7, respectively.

As mentioned above, we observe that subtle artifacts tend to be preserved by textural features from shallow layers of the network, thus we choose L2 and L3 as the candidates of $SL_t$. Conversely, we want the attention maps to attend to different regions of the input, which needs the guidance of high-level semantic information to some extent. Therefore, we use deeper stage L4 and L5 as the candidates of $SL_a$. By default setting $M=1$, we train models with four combinations on FF++(HQ). From the results in \Tref{tab_sl_selection}, we find that the model reaches best performance when using L2 for $SL_t$ and L5 for $SL_a$.

\begin{table}[t]
\centering

\setlength{\tabcolsep}{1.9mm}{
\begin{tabular}{c|c|c|c|c}
\toprule
\multirow{2}{*}{Methods} & \multicolumn{2}{c|}{LQ} & \multicolumn{2}{c}{HQ}          \\ \cline{2-5}
                     & ACC            & AUC           & ACC          & AUC            \\ \midrule
Steg.Features\cite{stegfeature}         & 55.98          & -                & 70.97         &                  \\ 
LD-CNN\cite{ldcnn}                & 58.69          & -                & 78.45          & -                \\ 
MesoNet\cite{mesonet}               & 70.47          & -                & 83.10          & -                \\ 
Face X-ray\cite{li2020face}            & -                & 61.60          & -                & 87.40          \\ 
Xception\cite{xception}             & 86.86          & 89.30          & 95.73          & 96.30          \\ 
Xception-ELA\cite{Gunawan2017DevelopmentOP}          & 79.63          & 82.90          & 93.86          & 94.80          \\
Xception-PAFilters\cite{Chen2017JPEGPhaseAwareCN}    & 87.16          & 90.20         & -                & -                \\ 
$\text{F}^3\text{-Net}$\cite{shao2020thinking}     & \textbf{90.43}  & \textbf{93.30}          & 97.52          & 98.10  \\
Two Branch\cite{masi2020two}           & -                & 86.59          & -                & 98.70          \\ 
EfficientNet-B4\cite{efficientnet}       & 86.67          & 88.20          & 96.63          & 99.18          \\\midrule
Ours(Xception)        & 86.95          & 87.26          & 96.37          & 98.97          \\ 
Ours(Efficient-B4)    & 88.69 & 90.40 & \textbf{97.60} & \textbf{99.29} \\ 
\bottomrule
\end{tabular}
}
 \vspace{-0.8em}  
\caption{Quantitative comparison on FaceForensics++ dataset with High-Quality and Low-Quality settings, respectively. The best performances are marked as bold.}
\label{tab_FF++}
\end{table}

\subsection{Comparison with Previous Methods}\label{sec_4.3}
In this section, we compare our framework with current state-of-the-art deepfake detection methods. We evaluate the performance on FF++ \cite{rossler2019faceforensics++} and DFDC \cite{dolhansky2020deepfake}, respectively. And we further evaluate the cross-dataset performance on Celeb-DF \cite{li2020celeb} in \Sref{tab_celeb}. We adopt ACC (accuracy) and AUC (area under Receiver Operating Characteristic Curve) as the evaluation metrics for extensive experiments.

\subsubsection{Evaluation on FaceForensics++}\label{sec_4.3.1}
FaceForensics++\cite{rossler2019faceforensics++} is the most widely used dataset in many deepfake detection approaches, it contains 1000 original real videos from internet and each real video corresponds to 4 forgery ones, which are manipulated by Deepfakes, NeuralTextures\cite{Neuraltexture}, FaceSwap\cite{Faceswap} and Face2Face\cite{thies2016face2face:}, respectively. In the training process, we augment the original frames 4 times for real/fake label balance. We adopt EfficientNet-B4 as the backbone of our framework, and test the performances on HQ (c23) version and LQ (c40) version, respectively. Specially, when training our model on LQ, the parameters are initialized by those pretrained on HQ to accelerate the convergence. The comparison results are listed in \Tref{tab_FF++}.

The results in \Tref{tab_FF++} demonstrate that our method achieves state-of-the-art performance on the HQ version of FF++. And the performances of different backbone verifies that our framework is not restricted by the backbone networks. However, the performance decreases 1.5\% compared with $\text{F}^3\text{-Net}$ \cite{shao2020thinking} on the LQ version since $\text{F}^3\text{-Net}$ is a specifically designed method for high-compressed deepfake videos detection. This is mainly because the videos in FF++(LQ) are highly compressed and cause a significant loss in textural information, which is a disaster to our texture enhancement designing. The results also reveal a limitation of our framework, that is, our framework is sensitive to high compression rate which blurs most of the useful information in spatial domain. We will make our framework more robust to compression in the future.

\subsubsection{Evaluating on DFDC Dataset}\label{sec_4.3.2}
DeepFake Detection Challenge (DFDC) is the most recently released largest scale deepfake detection dataset, this dataset is public on the Deepfake Detection Challenge organized by Facebook in 2020. Currently, it is the most challenging dataset for deepfake detection task due to the excellent forgery quality of fake videos in this dataset. Seldom previous methods have been conducted on this dataset thus we train our model on the training set of this dataset and only compare the logloss score with the winning teams' methods of the DFDC contest. Here the provided logloss scores are calculated on the DFDC testing set(Ref. to Table 2 of \cite{dolhansky2020deepfake}), which is one part of DFDC private set. Smaller logloss represent a better performance. The results in \Tref{tab_dfdc} demonstrate that our framework achieves state-of-the-art performance on DFDC dataset.

\begin{table}[t]
\centering
\setlength{\tabcolsep}{10.35mm}{
\begin{tabular}{c|c}
\toprule
Method    & Logloss          \\ \midrule
Selim  Seferbekov\cite{Seferbekov2020} & 0.1983           \\ 
WM\cite{wm2020}                 & 0.1787           \\ 
NTechLab\cite{NTechLab2020}           & 0.1703           \\ 
Eighteen Years Old\cite{Eighteen2020} & 0.1882           \\ 
The Medics\cite{Medics2020}         & 0.2157           \\ \midrule
Ours              & \textbf{0.1679}  \\
\bottomrule
\end{tabular}
}
\vspace{-0.8em}
\caption{Comparison with DFDC winning teams' methods on the DFDC testing dataset. We participated in the competition as team WM.}
\label{tab_dfdc}
\end{table}

\subsubsection{Cross-dataset Evaluation on Celeb-DF}\label{sec_4.3.3}
In this part, we evaluate the transferability of our framework, that is trained on FF++(HQ) with multiple forgery methods but tested on Celeb-DF \cite{li2020celeb}. We sample 30 frames for each video to calculate the frame-level AUC scores. The results are shown in \Tref{tab_celeb}. Our method shows better transferability than most existing methods. Two-branch \cite{masi2020two} achieves the state-of-the-art performance in transferability, however, its in-dataset AUC is far behind ours.

\begin{table}[t]
\centering
\setlength{\tabcolsep}{5.35mm}{
\begin{tabular}{c|c|c}
\toprule
Method         & FF++  & Celeb-DF \\ \midrule
Two-stream\cite{Zhou2017TwoStreamNN}     & 70.10 & 53.80    \\ 
Meso4\cite{mesonet}          & 84.70 & 54.80    \\ 
Mesolnception4\cite{mesonet} & 83.00 & 53.60    \\ 
FWA\cite{Li2019ExposingDV}           & 80.10 & 56.90    \\ 
Xception-raw\cite{li2020celeb}   & 99.70 & 48.20    \\ 
Xception-c23\cite{li2020celeb}   & 99.70 & 65.30    \\ 
Xception-c40\cite{li2020celeb}   & 95.50 & 65.50    \\ 
Multi-task\cite{Nguyen2019MultitaskLF}     & 76.30 & 54.30    \\
Capsule\cite{Nguyen2019CapsuleforensicsUC}        & 96.60 & 57.50    \\
DSP-FWA\cite{Li2019ExposingDV}       & 93.00 & 64.60    \\ 
Two Branch\cite{masi2020two}     & 93.18 & 73.41    \\ 
$\text{F}^3\text{-Net}$\cite{shao2020thinking}  & 98.10 & 65.17  \\
EfficientNet-B4\cite{efficientnet}     & 99.70 & 64.29    \\ \midrule
Ours           & 99.80 & 67.44    \\ 
\bottomrule
\end{tabular}
}
\vspace{-0.8em}
\caption{Cross-dataset evaluation on Celeb-DF (AUC(\%)) by training on FF++. Results of some other methods are cited directly from \cite{masi2020two}. Our method outperforms most deepfake detection approaches.}
\label{tab_celeb}
\end{table}

\subsection{Ablation Study}\label{sec_4.4}
\subsubsection{Effectiveness of Multiple Attentions}\label{sec_4.4.1}

To confirm the effectiveness of using multiple attentions, we evaluate how the quantity of attention maps affect the accuracy and transferability of our model. We train models in our framework with different attention quantities $M$ on FF++(HQ), the other hyper-parameters are kept same as settings in \Tref{tab_FF++}. For the single attentional model, we do not use the regional independence loss and AGDA. 
\begin{table}[t]
\centering
\setlength{\tabcolsep}{8.3mm}{
\begin{tabular}{c|c|c}
\toprule
M &  FF++(HQ) & Celeb-DF\\ \midrule
1 & 97.26          & 67.30          \\
2 & 97.51          & 65.74              \\
3 & 97.35          & 66.86              \\
4 & \textbf{97.60} &  \textbf{67.44}    \\ 
5 & 97.39          & 66.82               \\ 
\bottomrule
\end{tabular}
}
\vspace{-0.8em}
\caption{Ablation results on FF++(HQ) (Acc \%) and Celeb-DF (AUC \%) with different number of attention maps.}
\label{tab_num_attention}
\end{table}

The Acc results on FF++(HQ) and AUC results on Celeb-DF are reported in table \Tref{tab_num_attention}. In some cases, multi-attention based models perform better than the single attentional model, and we found that $M=4$ provides the best performance.

\begin{figure}[t]
    \begin{center}
        \includegraphics[width=1\linewidth]{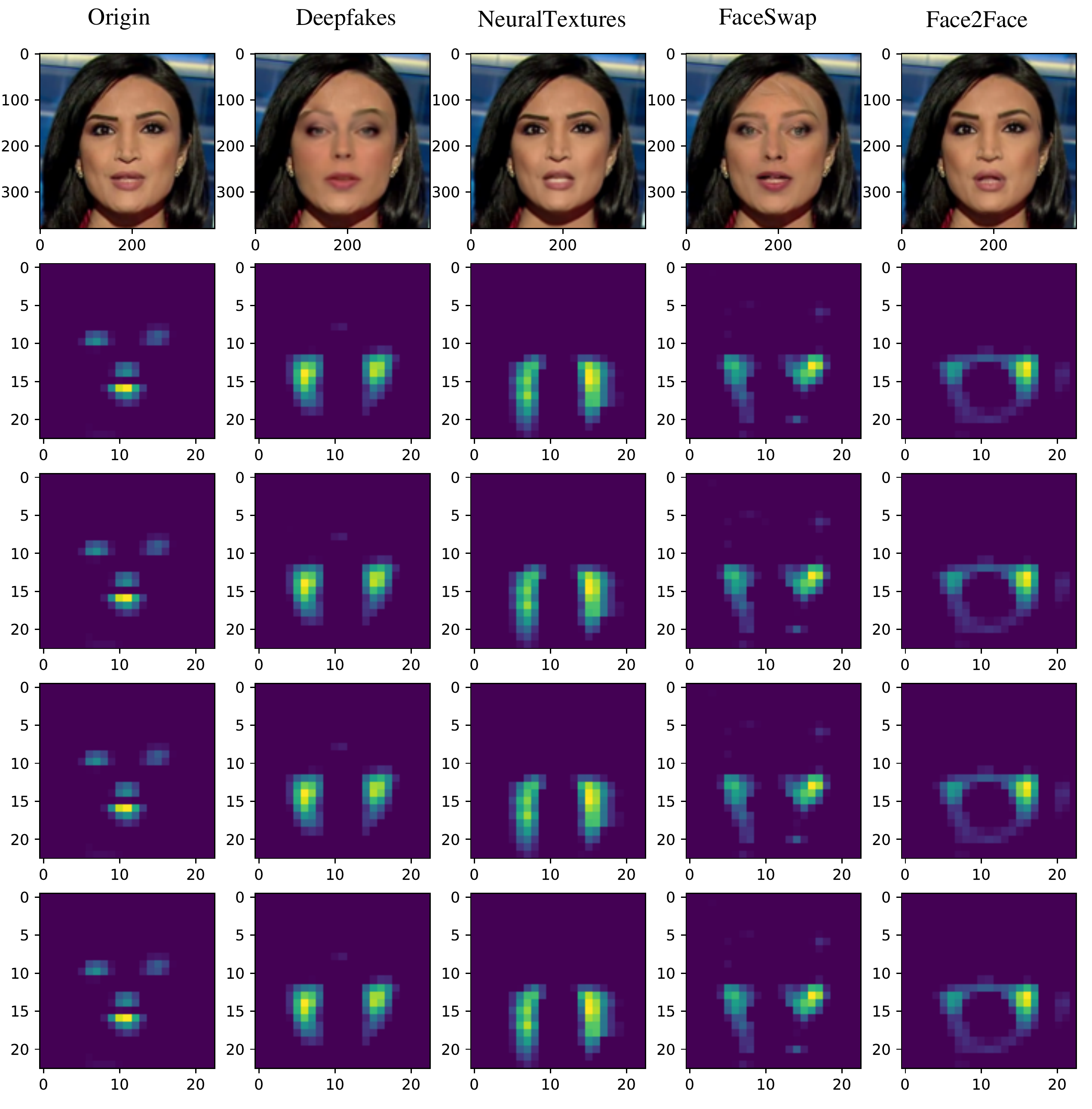}
    \end{center}
    \vspace{-0.8em}
    \caption{Attention maps trained without regional independence loss (RIL) and AGDA. Without RIL and AGDA, the network is easily degraded and the multiple attention maps locates the same regions of input.}
    \label{fig_bad_attention_maps}
\end{figure}

\begin{figure}[t]
    \begin{center}
        \includegraphics[width=1\linewidth]{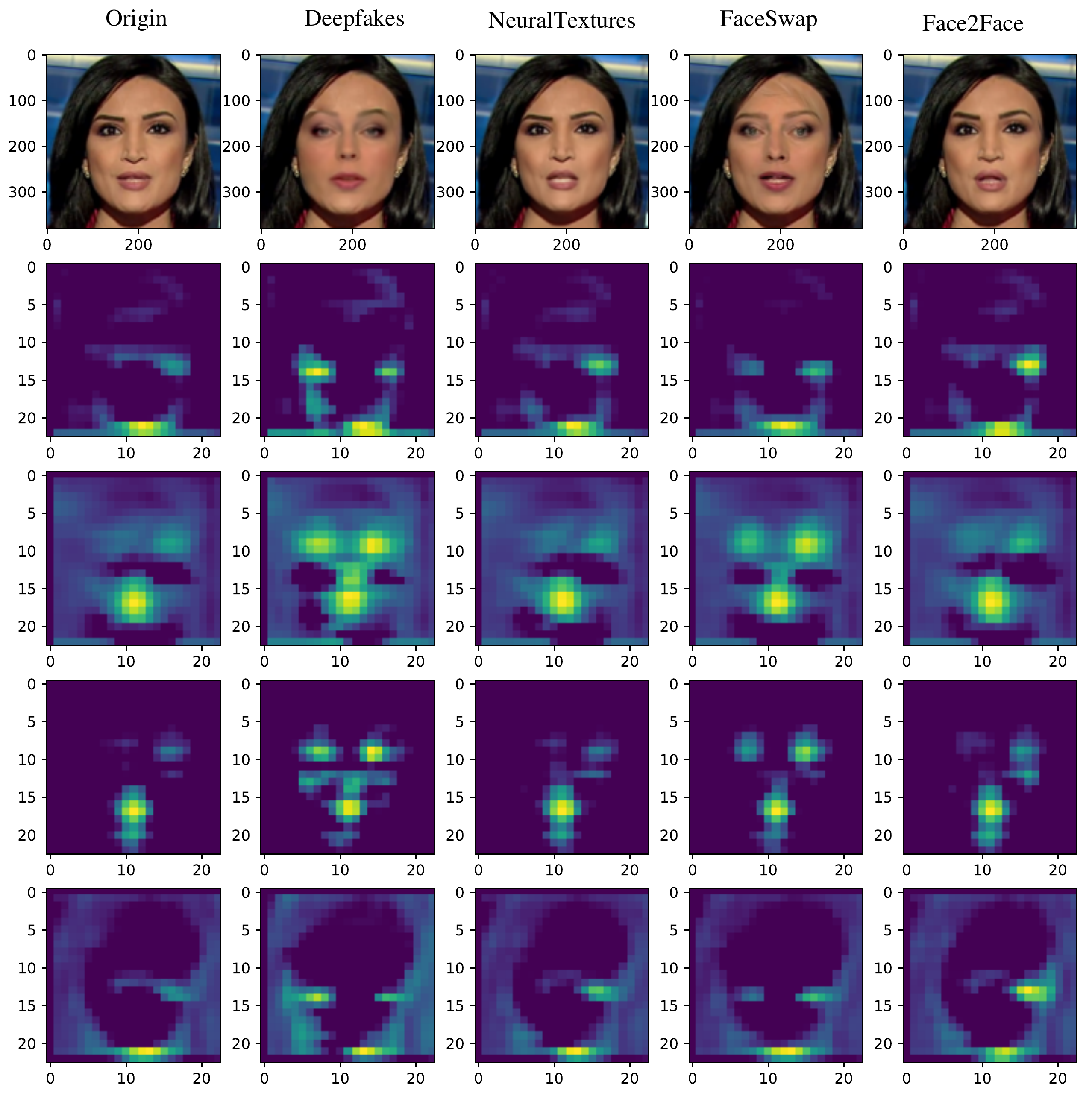}
    \end{center}
    \vspace{-0.8em}
    \caption{Attention maps trained without AGDA. Although the regional independence loss forces different attention maps to separate, they tend to respond to the same salient feature without the help of AGDA.}
    \label{fig_overlapattentions}
\end{figure}

\subsubsection{Ablation Study on Regional Independence Loss and AGDA}\label{sec_4.4.2}
As mentioned above, the regional independence loss and AGDA play an important role in regularized multiple attention maps training. In this part, we execute quantitative experiments and give some visualizations to demonstrate that these two components are necessary.

First, to demonstrate the effectiveness of our regional independence loss, we compare the performances of the models trained with different auxiliary losses. We keep all the settings identical with previous except for the loss function. With the same motivation in designing auxiliary loss, we substitute the regional independence loss with Additive Angular Margin softmax(AMS)\cite{amsoftmax} that can also force feature vectors close to their center.

Then we verify the effectiveness of our design for AGDA. As mentioned, we blur the original image to degrade the selected region of input. Thus the strategy of AGDA can be regarded as a ``soft attention dropping". In this part, we alternatively adopt a ``hard attention dropping", which directly erases pixels of selected region by binary attention mask $BM$:

\begin{equation}
  \begin{array}{l}
        BM_k(i, j)=\begin{cases}
            0, & \text{if $A^*_k(i, j) > \theta_d $}\\
            1, & \text{otherwise}.
          \end{cases} \\
  \end{array}
    \label{eq_binary_mask}
\end{equation}
We set the attention dropping threshold $\theta_d=0.5$ in this experiment. The comparison results of this ablation study are depicted in \Tref{tab_loss_agda}. The results verify that both regional independence loss (RIL) and attention guided data augmentations (soft attention dropping) have remarkable contribution to improve the performance of our framework.

\begin{figure}[t]
    \begin{center}
        \includegraphics[width=1\linewidth]{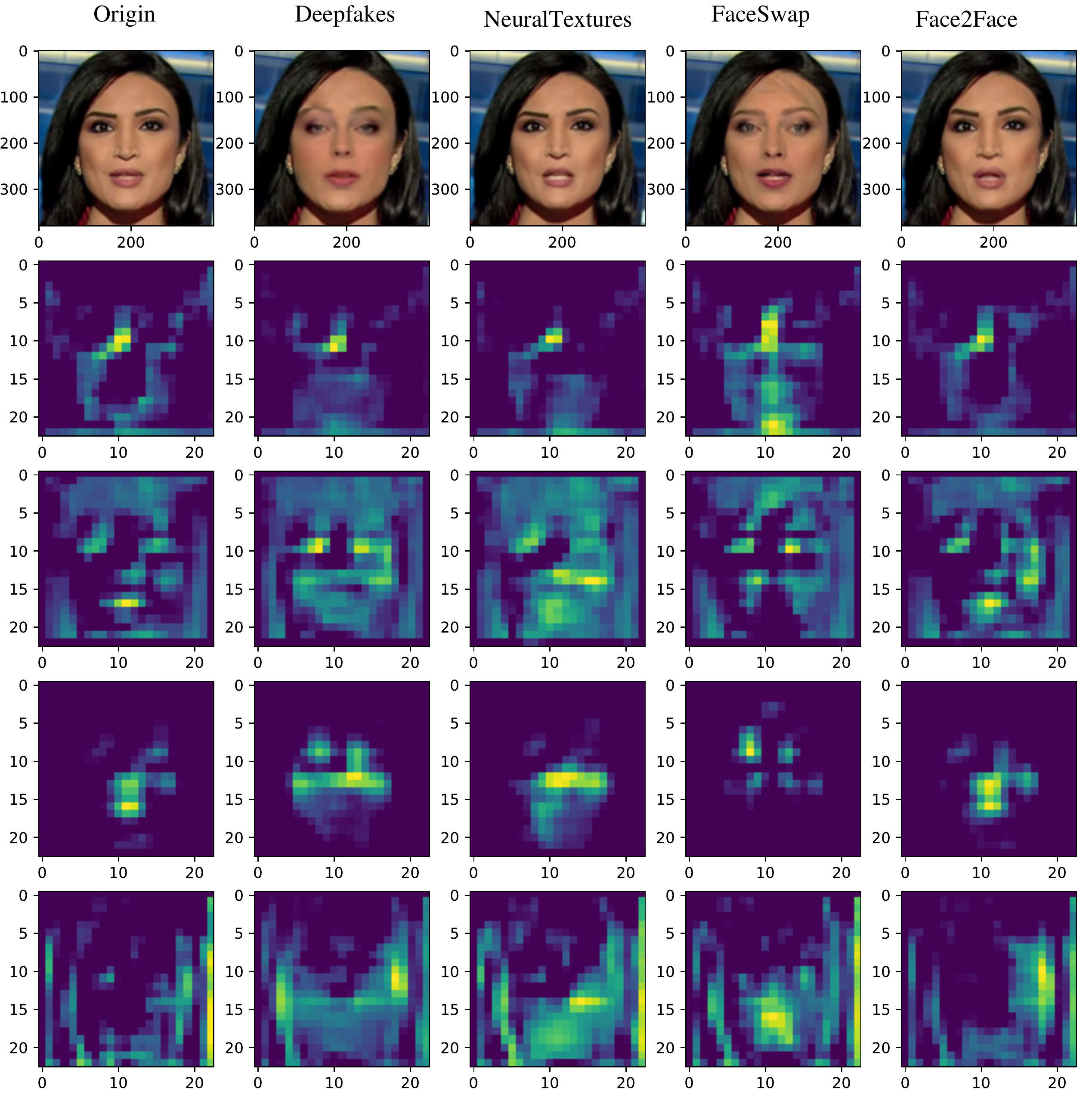}
    \end{center}
    \vspace{-0.8em}
    \caption{Attention maps trained with both regional independence loss and AGDA. The location and response of attention maps are correctly distributed. }
    \label{fig_good_example}
    \vspace{-0.5em}
\end{figure}

\begin{table}[t]  
\centering
\setlength{\tabcolsep}{3mm}{
\begin{tabular}{c|c|c|c}
\toprule
Loss type  & AGDA type     & FF++(HQ) & Celeb-DF\\ \midrule
None          & None          & 96.74        & 64.86               \\ 
AMS           & None          & 96.49        & 64.23               \\ 
RIL          & None          &  97.38        & 65.85               \\
AMS          & Hard          &  96.53        & 63.73               \\ 
RIL           & Hard          &  97.24        & 64.40               \\ 
AMS           & Soft          &  96.78        & 66.42               \\ 
RIL  &  Soft & \textbf{97.60} & \textbf{67.44}     \\ \hline
\end{tabular}
}
\vspace{-0.8em}
\caption{Ablation results of different loss function and AGDA strategy. The model achieves best performance when using regional independence loss and soft AGDA mechanism. The metric on FF++(HQ) dataset is ACC, and on Celeb-DF is AUC.}
\label{tab_loss_agda}
\end{table}

To further help understanding of the function of regional independence loss and the AGDA strategy, we visualize the attention maps of models trained with/without these two components. \Fref{fig_bad_attention_maps} illustrate the attention maps without RIL, it shows a clear trend that all attention maps are focused on same region. \Fref{fig_overlapattentions} demonstrate that, although the attention regions are separated under the retraining of RIL, the different regions still exhibit similar response to the most salient features such as landmarks. This is not conductive for multiple attention maps to capture divergent information from different regions. While \Fref{fig_good_example} verifies that when both RIL and soft AGDA are adopted, the attention maps show response in discriminative regions with diverse semantic representations.

\section{Conclusion}\label{sec_5}

In this paper, we research the deepfake detection from a novel perspective that is formulating the deepfake detection task as a fine-grained classification problem. We propose a multi-attentional deepfake detection framework. The proposed framework explores discriminative local regions by multiple attention maps, and enhances texture features from shallow layers to capture more subtle artifacts. Then the low-level textural feature and high-level semantic features are aggregated guided by the attention maps. A regional independence loss function and attention guided data augmentation mechanism are introduced to help train disentangled multiple attentions. Our method achieves good improvements in extensive metrics.

{\small
\bibliographystyle{ieee_fullname}
\bibliography{egpaper_for_review}
}

\end{document}